\documentclass[doubleblind]{ecai}

\usepackage{amsmath,amsfonts,bm}

\newcommand{\ie}{{\textit{i.e.}}}

\def\Figref#1{Figure~\ref{#1}}

\def\twoFigref#1#2{Figures \ref{#1} and \ref{#2}}

\def\eqref#1{~eq.(\ref{#1})}
\def\Eqref#1{~Eq.(\ref{#1})}

\def\1{\bm{1}}

\DeclareMathAlphabet{\mathsfit}{\encodingdefault}{\sfdefault}{m}{sl}
\SetMathAlphabet{\mathsfit}{bold}{\encodingdefault}{\sfdefault}{bx}{n}

\newtheorem{definition}{Definition}
\usepackage[ruled,vlined]{algorithm2e}
\usepackage{multirow}
\usepackage{enumitem}
\usepackage{graphicx}
\usepackage{amsmath}
\usepackage{amssymb}
\usepackage{gensymb}
\usepackage{booktabs}

\usepackage{makecell}
\usepackage{colortbl}
\usepackage{xcolor}
\usepackage{latexsym}

\usepackage{hyperref}
\hypersetup{pagebackref,breaklinks,colorlinks}

\begin{document}

\begin{frontmatter}

\title{DiffuseGAE: Controllable and High-fidelity Image Manipulation from Disentangled Representation}
\author{{Yipeng}~{Leng}}
\author{\fnms{Qiangjuan}~\snm{Huang}$^\ast$}
\author{\fnms{Zhiyuan}~\snm{Wang}$^\ast$}
\author{\fnms{Yangyang}~\snm{Liu}$^\ast$}
\author{\fnms{Haoyu}~\snm{Zhang}$^\ast$}
\address{National Innovation Institute of Defense Technology}
\thanks{Corresponding Author.}


\begin{abstract}
Diffusion probabilistic models (DPMs) have shown remarkable results on various image synthesis tasks such as text-to-image generation and image inpainting. However, compared to other generative methods like VAEs and GANs, DPMs lack a low-dimensional, interpretable, and well-decoupled latent code. Recently, diffusion autoencoders (Diff-AE) were proposed to explore the potential of DPMs for representation learning via autoencoding. Diff-AE provides an accessible latent space that exhibits remarkable interpretability, allowing us to manipulate image attributes based on latent codes from the space. However, previous works are not generic as they only operated on a few limited attributes. To further explore the latent space of Diff-AE and achieve a generic editing pipeline, we proposed a module called \textbf{G}roup-supervised \textbf{A}uto\textbf{E}ncoder(dubbed GAE) for Diff-AE to achieve better disentanglement on the latent code. Our proposed GAE has trained via an \textbf{attribute-swap} strategy to acquire the latent codes for multi-attribute image manipulation based on examples. We empirically demonstrate that our method enables multiple-attributes manipulation and achieves convincing sample quality and attribute alignments, while significantly reducing computational requirements compared to pixel-based approaches for representational decoupling. Code will be released soon.
\end{abstract}

\end{frontmatter}


\section{Introduction}
Score-based generative models (SGMs)~\cite{song2020score, vahdat2021score} and Diffusion probabilistic model (DPMs)~\cite{ho2020denoising, song2020denoising} have gained great attention for its training stability, scalability, and impressive image synthesis quality. These models have been shown to achieve impressive performance on diverse domains spanning computer vision~\cite{amit2021segdiff, li2022srdiff, luo2021score}, natural language processing~\cite{li2022diffusion, yu2022latent}, multi-model modeling~\cite{bao2023unidiffuser, saharia2022photorealistic}. Nevertheless, the diffusion model has not been well investigated as of yet.

The approach of unsupervised representation learning via generative modeling is currently a highly hot issue. Some latent variable generative models, like GANs and VAEs, are inherently candidates for them. We can easily get interpretable latent codes in the process of generative modeling. But in diffusion models, we only get a sequence of pixel-level noise and intermediate reconstructed images which lack high-level semantic information. In light of this,  Diff-AE~\cite{pandey2022diffusevae, preechakul2022diffusion, zhang2022unsupervised} is an insightful investigation of how to use conditional diffusion model for representation learning via autoencoding. Specifically, they use a traditional 
CNN encoder~$\boldsymbol{E}_{\phi}$ for digging meaningful representations $\textbf{z}_{sem}$ from images, then employ a conditional DPM as the decoder~$\boldsymbol{D}_{\theta}$ for image reconstruction, taking latent code $\bm{z}_{sem} \in \mathcal{Z}$ as input of it. Following Ho et al.~\cite{ho2020denoising}, training is achieved  by optimizing $\mathcal{L}_\mathrm{simple}$ loss function with respect to $\phi$ and $\theta$.
\begin{equation}\label{equ:l_simple}
    \mathcal{L}_{\mathrm{simple}}:=\mathbb{E}_{t, \mathbf{x}_0, \boldsymbol{\epsilon}}\left[\left\|\boldsymbol{\epsilon}\!-\!\boldsymbol{\epsilon}_\theta\left(\sqrt{\bar{\alpha}_t} \mathbf{x}_0\!+\!\sqrt{1\!-\!\bar{\alpha}_t} \boldsymbol{\epsilon}, t, \textbf{z}_{sem}\right)\right\|^{2}_{2}\right]
\end{equation}
where $\boldsymbol{\epsilon} \in \mathbb{R} ^ {3\times h \times w} \sim \mathcal{N} (0, I)$, $\mathbf{x}_0$ is input image, $\bm{z}={\boldsymbol{E}}_{\phi}(\mathbf{x}_0)$, $\bar{\alpha}_{t}$ is a hyperparameter. Diff-AE is competitive with state-of-the-art models on various datasets' benchmarks and downstream tasks. More importantly, we can utilize its semantic-rich latent space to perform image synthesis. 

Latent variable generative models, such as VAEs and GANs, are imposed independent and identical constraints on their prior distribution, making them inherently decoupled. 
Yet, Diff-AE is not bounded by such restraints, there is still significant promise in the realm of controllable image synthesis, like multi-attribute image manipulation and precise image inpainting.
To address the aforementioned issues, we further explore the disentanglement capability of Diff-AE semantic space $\mathcal{Z}$.
We can obtain a more meaningful and readily controllable latent code $\textbf{z}_{sem}\in\mathcal{W}$ and utilize it to perform image synthesis.
$\mathcal{Z} ~\text{and} ~\mathcal{W}$ respectively denote latent space obtained by initial decoupling and further decoupling.
To this end, we propose a novel decoupled representation learning framework called group-supervised diffusion autoencoder (dubbed DiffuseGAE) for zero-shot image synthesis by recombination of multi-attribute latent code.
As shown in Fig.~\ref{fig:pipeline}, DiffuseGAE consists of a diffusion autoencoder and a novel autoencoder called group-supervised autoencoder(GAE).
It is known that diffusion models are often criticized for their long generation time, so we adopt conditional Denoising Diffusion Implicit Model(DDIM) which can be transformed into deterministic Diffusion. It is the first-order case of DPM-solver, a fast sampling method that makes full use of the semi-linearity of diffusion ODEs. Under these circumstances, we can easily transfer it into a more efficient method. 
In addition, the determinism of DDIM allows us to obtain a consistent mapping between the image and the noise, which guarantees the quality of image reconstruction.

Following the paradigm of Diff-AE~\cite{preechakul2022diffusion}, a diffusion autoencoder composed of a CNN encoder and the aforementioned DDIM is trained in an end-to-end manner.
To more fully exploit the latent space encoded by the above autoencoder, we proposed a module called \textbf{G}roup-Supervised \textbf{A}uto\textbf{E}ncoder(GAE) trained in latent space. It is used for decoupling latent space encoded by the trained diffusion autoencoder. We train it in an attribute-swap strategy mentioned in GZS-Net~\cite{GeAXI21}. The latent code generated by GAE is passed to DDIM as new $\bm{z}_{sem}$ for image generation. But unlike GZS-Net, we impose constraints on the latent space, making GAE an efficient fine-tuning of the original diffusion autoencoder.
In order to narrow the gap between the inputs of the prior and subsequent DDIM, we then jointly train the two previously mentioned models in an end-to-end manner.

 We conduct several experiments which clearly demonstrate the effectiveness and generalization of our proposed framework. We can obtain a more decoupled, controllable, and intuitive representation space through our method. We qualitatively and quantitatively demonstrate that using our approach can achieve better performance on information compression and representation disentanglement. Thanks to the better-decoupled latent space, our quality in multi-attribute image reconstruction, and image interpolation are competitive with previous state-of-the-art decoupled representation models. 
 
 Our main contributions can be summarized as follows:
\begin{itemize}
\item We further explore the disentanglement capability of the latent code extracted from the diffusion autoencoder and enable each part of the latent code to represent an attribute.
  
\item We design a plug-in module called Group-supervised AutoEncoder(\textbf{GAE}) for disentanglement and precise reconstruction in latent codes and images. 

\item We propose \textbf{DiffuseGAE}, a novel decoupled representation learning framework capable of better image reconstruction and multi-attribute disentanglement, as well as zero-shot image synthesis than other techniques for representation decoupling.
\end{itemize}

\section{Background}
\label{sec:bg}
\vspace{-1pt}
\subsection{Denoising Diffusion Probabilistic Models}
DDPM\cite{ho2020denoising} is modeled as $\boldsymbol{x}=\boldsymbol{x}_0 \rightleftharpoons \boldsymbol{x}_1 \rightleftharpoons \boldsymbol{x}_2 \rightleftharpoons \cdots \rightleftharpoons \boldsymbol{x}_{T-1} \rightleftharpoons \boldsymbol{x}_T=\boldsymbol{n}$. The forward process $q(x_t|x_{t-1})$ is transforming data $x$ into random noise $n$ in a Markovian manner and the reverse process $p(x_{t-1}|x_{t})$ is restoring $n$ to $x$, which is the generative model we desired. 
Unlike VAEs, the forward process $q$ is modeled as a constant normal distribution.
\vspace{-4pt}
\begin{equation}
    \begin{aligned}
    q\left(\boldsymbol{x}_t | \boldsymbol{x}_{t-1}\right)&=\mathcal{N}\left(\boldsymbol{x}_t ; \alpha_t \boldsymbol{x}_{t-1}, \beta_t^2 \boldsymbol{I}\right)\quad \\
    q\left(\boldsymbol{x}_t | \boldsymbol{x}_0\right)&=\mathcal{N}\left(\boldsymbol{x}_t ; \bar{\alpha}_t \boldsymbol{x}_0, \bar{\beta}_t^2 \boldsymbol{I}\right)\quad
    \end{aligned}
\end{equation}
where $\alpha_t^2+\beta_t^2=1$,~$\bar{\alpha}_t=\alpha_1\alpha_2 \cdots \alpha_t $, $\bar{\beta_t} = \sqrt{1-\bar{a_t}^{2}}$. We can get $q(x_{t-1}|x_t, x_0)$ using Bayes theorem.
\begin{equation}
    \begin{aligned}
    q(\boldsymbol{x}_{t\!-\!1}|\boldsymbol{x}_t, &\boldsymbol{x}_0)\!
    =\frac{q\left(\boldsymbol{x}_t \mid \boldsymbol{x}_{t-1}\right) q\left(\boldsymbol{x}_{t-1} \mid \boldsymbol{x}_0\right)}{q\left(\boldsymbol{x}_t \mid \boldsymbol{x}_0\right)}\\
    &=\!\mathcal{N}\left(\boldsymbol{x}_{t\!-\!1} ; \frac{\alpha_t \bar{\beta}_{t\!-\!1}^2}{\bar{\beta}_t^2} \boldsymbol{x}_t\!+\!\frac{\bar{\alpha}_{t\!-\!1} \beta_t^2}{\bar{\beta}_t^2} \boldsymbol{x}_0, \frac{\bar{\beta}_{t\!-\!1}^2 \beta_t^2}{\bar{\beta}_t^2} \boldsymbol{I}\right)
    \end{aligned}
\end{equation}
It is straightforward to get the variational lower bound using Jensen’s inequality when trading the cross entropy $\mathcal{L}_{CE}$ as optimization objectives, and $\mathcal{L}_{\text{VLB}}$ can be further rewritten to a combination of KL-divergence terms. 
\vspace{-5pt}
\begin{equation}\label{equ:loss_ce}
\begin{aligned}
\mathcal{L}_{\text{CE}}&=-\mathbb{E}_{q\left(\mathbf{x}_0\right)} \log p_\theta\left(\mathbf{x}_0\right) \\
& \leq\mathbb{E}_{q\left(\mathbf{x}_{0: T)}\right.}\left[\log \frac{q\left(\mathbf{x}_{1: T} | \mathbf{x}_0\right)}{p_\theta\left(\mathbf{x}_{0: T}\right)}\right]=L_{\text{VLB}} \\
& =L_T+L_{T-1}+\cdots+L_0
\end{aligned}
\end{equation}
where $L_T =D_{\text{KL}}\left[q\left(\mathbf{x}_T|\mathbf{x}_0\right) \| p_\theta\left(\mathbf{x}_T\right)\right]$, 
$L_t =D_{\text{KL}}[q\left(\mathbf{x}_t | \mathbf{x}_{t+1}, \mathbf{x}_0\right)$
$\| p_\theta\left(\mathbf{x}_t | \mathbf{x}_{t+1}\right)]$
~~$\text{for 1} \leq t \leq T-1$. $L_T$ will be a constant because there are no training parameters in the forward process$q$, the loss term $L_t$ is parameterized to minimize difference between real noise and predicted noise. Ho et al.\cite{ho2020denoising} empirically found that diffusion models similar to equation \ref{equ:l_simple} work better when ignoring the weight term. After training, We can iteratively generate images via $p(x_{t-1} | x_{t})$ from random noise.
\vspace{-2pt}
\begin{equation}
\begin{aligned}
\mathcal{L}_t & =\mathbb{E}_{\mathbf{x}_{0, \boldsymbol{\epsilon}}}\left[\frac{1}{\left\|\mathbf{\sigma}_\theta\left(\mathbf{x}_t, t\right)\right\|_2^2}\left\|\tilde{\boldsymbol{\mu}}_t\left(\mathbf{x}_t, \mathbf{x}_0\right)-\boldsymbol{\mu}_\theta\left(\mathbf{x}_t, t\right)\right\|^2\right] \\
& =\mathbb{E}_{\mathbf{x}_{0, \epsilon} \boldsymbol{\epsilon}}\left[\frac{\left(1-\alpha_t\right)^2}{\alpha_t\left(1-\bar{\alpha}_t\right)\left\|\mathbf{\sigma}_\theta\right\|_2^2}\left\|\boldsymbol{\epsilon}_t-\boldsymbol{\epsilon}_\theta\left(\mathbf{x}_t, t\right)\right\|^2\right] \\
&\propto \mathbb{E}_{t, \mathbf{x}_{0, \epsilon}, \epsilon_t}\left[\left\|\boldsymbol{\epsilon}_t-\boldsymbol{\epsilon}_\theta\left(\mathbf{x}_t, t\right)\right\|^2\right] = \mathcal{L}_{\text {simple}}\\
\end{aligned}
\end{equation}
\vspace{-15pt}

\subsection{Denoising Diffusion Implicit Models}
From the derivation of the DDPM mentioned above, it can be viewed that the estimation of $p(x_{t-1} | x_t)$ is in a progressive manner. We find that the loss function only depends on $p(x_t | x_0)$, while the sampling process only depends on $p(x_{t-1} | x_t)$, $p(x_t | x_{t-1})$ and $p(x_t | x_{t-1})$ does not appear in the whole process. In sight of this, Song et al.~\cite{song2020denoising} reparameterize the posterior $q(x_{1:T}|x_0)$, turning it into a non-markovian process. Since we have relaxed the condition $p(x_{t} | x_{t-1})$, $p(x_{t-1} | x_t, x_0)$ should have a larger solution space, \textit{i.e.} $\int p\left(\boldsymbol{x}_{t-1} | \boldsymbol{x}_t, \boldsymbol{x}_0\right) p\left(\boldsymbol{x}_t | \boldsymbol{x}_0\right) d \boldsymbol{x}_t=p\left(\boldsymbol{x}_{t-1} | \boldsymbol{x}_0\right)$. Song et al. derive $p(x_{t-1} | x_t, x_0)$ using undetermined coefficient method. We can replace $x_0$ with $\frac{1}{\bar{\alpha}_t}\left(\boldsymbol{x}_t-\bar{\beta}_t \boldsymbol{\epsilon}_{\boldsymbol{\theta}}\left(\boldsymbol{x}_t, t\right)\right)$ and reuse DDPM's parameter because ~$p(x_t|x_0)$ stayed unchanged, then get:
\begin{equation}
\label{equ:generate_process}
\begin{aligned}
    p(&\boldsymbol{x}_{t-1} | \boldsymbol{x}_t)  \approx p\left(\boldsymbol{x}_{t-1} | \boldsymbol{x}_t, x_0=\frac{1}{\bar{\alpha}_t}\left(\boldsymbol{x}_t-\bar{\beta}_t \boldsymbol{\epsilon}_{\boldsymbol{\theta}}\left(\boldsymbol{x}_t, t\right)\right)\right) \\
    & =\mathcal{N}( \frac{1}{\alpha_t}\left[\boldsymbol{x}_t\!-\!\left(\bar{\beta}_t\!-\!\alpha_t \sqrt{\bar{\beta}_{t-1}^2\!-\!\sigma_t^2}\right) \boldsymbol{\epsilon}_{\boldsymbol{\theta}}\left(\boldsymbol{x}_t, t\right)\right)],\sigma_t^2 \textbf{I} )
\end{aligned}
\end{equation}
When $\sigma_t = 0$,  they empirically observed better quality compared with other $\sigma$ settings. Meanwhile, the generative process $p$ becomes deterministic, which is a stable mapping from noise to reconstructed image. In addition, due to the semi-linear property of diffusion ODEs, DDIM has better scalability in terms of image generation acceleration. That is why we used DDIM for our decoder.

\section{Methodology}
\label{sec:method}
\subsection{Defects in the latent space of diffusion autoencoder}
\begin{figure*}
\centering
\vspace{-10pt}
\setlength{\abovecaptionskip}{-1pt}
\includegraphics[width=1.0\linewidth]{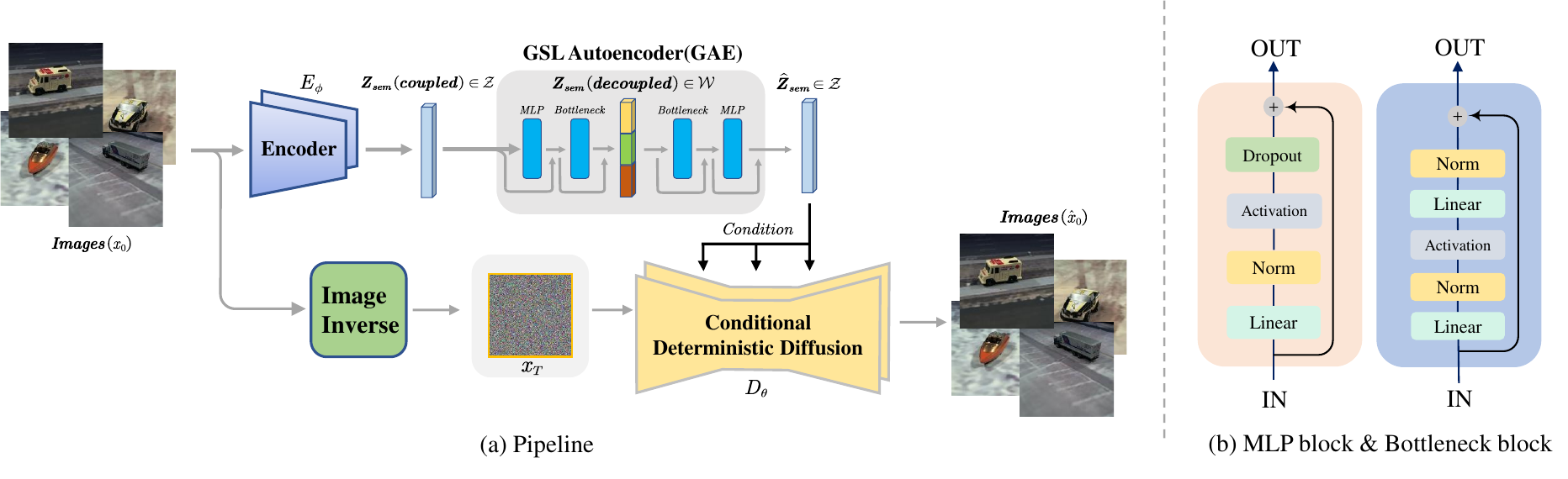}
\caption{\textbf{(a)}~\textbf{Illustration of our pipeline and dataflow.} Our proposed approach, named DiffuseGAE, consists of two main components: the diffusion autoencoder and the Group-supervised AutoEncoder (GAE). The Diff-AE is employed to initially capture the latent space $\mathcal{Z}$, which can be trained from scratch or using a pre-trained model. Subsequently, we use the GAE to further explore the hidden information and obtain $\mathcal{W}$. The reconstructed $\hat{z}_{sem}$ is passed as a condition to generate images using DDIM. Moreover, for more accurate reconstruction, we need to invert the images $x_0$ into a noise map $x_T$ and pass it to DDIM. \textbf{(b)} Illustrates the architecture of the MLP block and Bottleneck block used in GAE.}
\label{fig:pipeline}
\vspace{-3pt}
\end{figure*}

\begin{figure}[htb]
\centering
\includegraphics[width=0.9\linewidth]{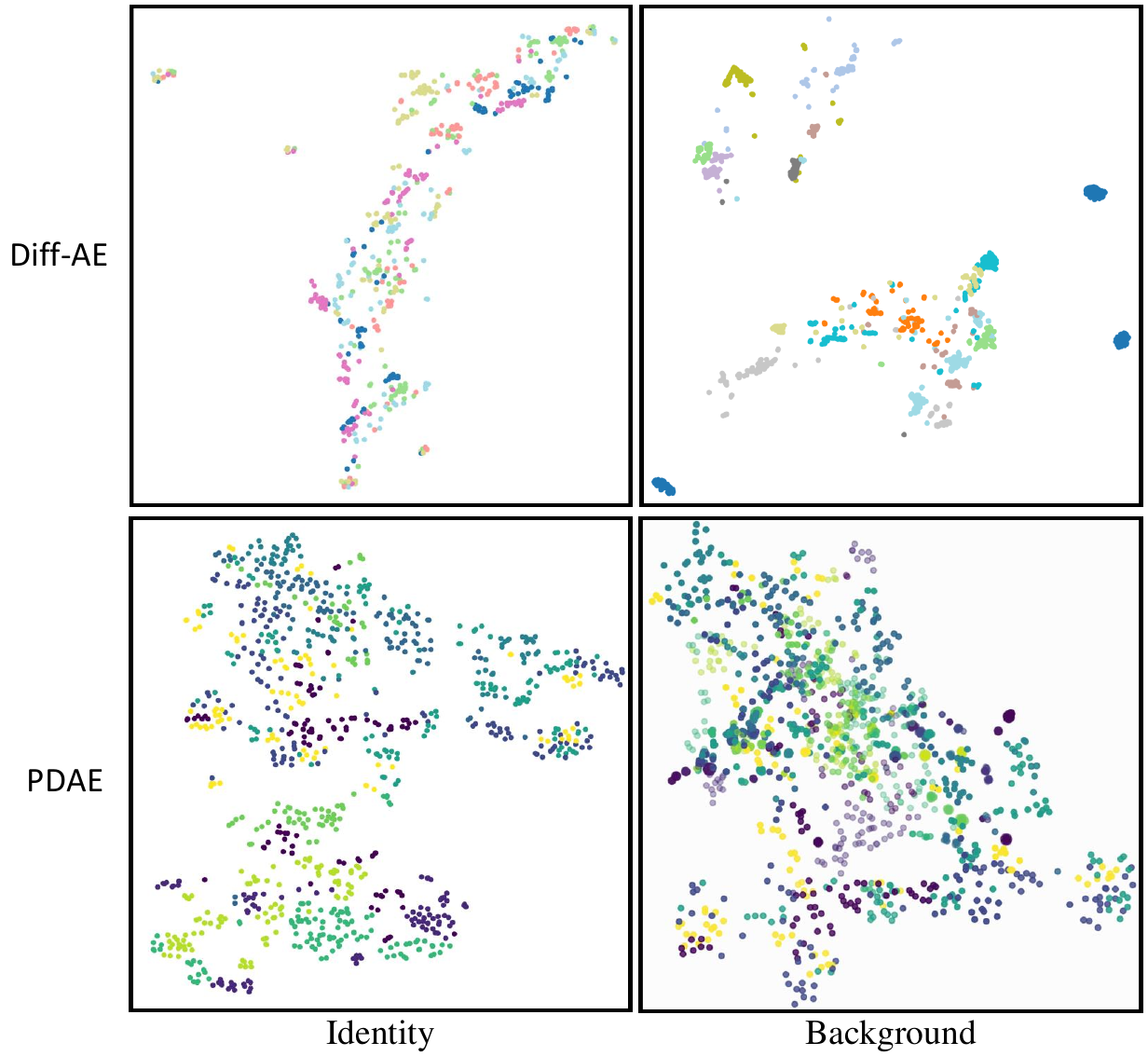}
\caption{Latent space of different attributes in Diff-AE~\cite{preechakul2022diffusion} and PDAE~\cite{zhang2022unsupervised}, we perform T-SNE visualization for identity(10 categories) and background(14 categories) using their latent code. Although the quality using diffusion autoencoder for image synthesis is excellent, their latent space interpretability is poor.}
\label{fig:movtivation}
\end{figure}
We are inspired by observations made regarding Diff-AE's latent space. As mentioned in \cite{pandey2022diffusevae,preechakul2022diffusion}, we can perform single-attribute image synthesis by manipulating the latent code because of its semantic and interpretable latent space. However, such methods are not effective when dealing with more fine-grained categories.
In Figure~\ref{fig:movtivation}, we perform T-SNE visualization with Diff-AE~\cite{preechakul2022diffusion} and PDAE~\cite{zhang2022unsupervised} on ilab\_20M dataset, a toy vehicles dataset for using
different image variations(e.g. identity, pose, background) in image synthesis. We observe that their latent space does not match their generative power, which hinders the further development of such models. Essentially, if we can further explore the latent space of diffusion autoencoder, for instance, by making it more decoupled and interpretable, we can effectively perform multi-attribute image synthesis. Moreover, according to Eq.(\ref{equ:loss_ce}), the more interpretable information of images~$x_0$ that~$\textbf{z}_{sem}$ contains, the smaller the mean gap exisits between $p(x_{t-1}|x_t)$ and $q(x_{t-1}|x_t)$ will be. 

Thus we propose a generic learning framework called DiffuseGAE which disentangles the space of diffusion autoencoder to achieve more controllable image manipulation.~\Figref{fig:pipeline} shows our pipeline and dataflow for DiffuseGAE. The training process of our proposed method consists of three steps, \ie ~pre-training, refinement, and fine-tuning. 

\subsection{Pre-training of DiffuseGAE}
In the pre-training process, following the paradigm of diffusion autoencoder~\cite{preechakul2022diffusion, zhang2022unsupervised}, we employ an encoder $\boldsymbol{E}_\phi(x_0)$ for getting latent space from input images and adopt a conditional deterministic diffusion(DDIM) to the decoder~$\boldsymbol{D_\theta}$ for image reconstruction. 

\begin{figure}[htb]
\centering
\vspace{-10pt}
\setlength{\abovecaptionskip}{-2pt}
\includegraphics[width=1.0\linewidth]{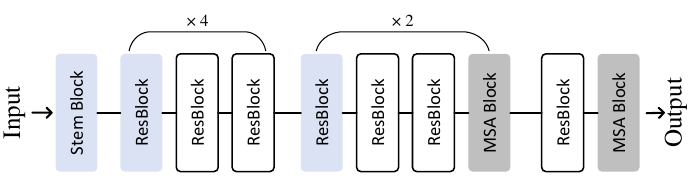}
\caption{Detailed architecture of Alter-ResNet. The architecture consists of Conv, Multi-head Self-Attention (MSA), and Downsample blocks represented by blue, white, and gray blocks respectively. To improve the quality of the latent space, we adopt Group Normalization  and SiLU for normalization and activation, respectively. }
\label{fig:encoder}
\end{figure}

\noindent \textbf{Encoder}
we follow the build-up rule in Park et al.~\cite{park2022vision} and redesign a new encoder called Alter-ResNet that replaces the conv block closer to the end of a stage with an attention block. The architecture of our Alter-ResNet is illustrated in Fig.~\ref{fig:encoder}.
In addition, as shown in Eq.(\ref{equ:reversed_process}), the diffusion model requires two entries: $\text{2-D noise code} ~x_t ~\text{and} ~z_{sem}$. To transform images into related noise codes, we develop a stochastic encoder by re-deriving the reversed process of DDIM and DPM-solver++~\cite{lu2022dpm, song2020denoising}. 
Since both approaches have the same posterior distribution~$q(x_t|x_0)$, their image inverse processes are equal, \ie


\begin{equation}
\label{equ:reversed_process}
\boldsymbol{x}_{t+1}=\sqrt{\bar{\alpha}_{t+1}}\hat{x}_0
+ \sqrt{1\!-\bar{\alpha}_{t+1}} \cdot {\boldsymbol{\epsilon}}_\theta\left(\bm{z},\boldsymbol{x}_t, t\right)
\end{equation}
where $\hat{x}_0 = \left({\boldsymbol{x}_t\!-\sqrt{1\!-\bar{\alpha}_t} \cdot {\boldsymbol{\epsilon}}_\theta\left(\bm{z},\boldsymbol{x}_t, t\right)}\right)/{\sqrt{\bar{\alpha}_t}}$.

\noindent \textbf{Decoder}
As previously noted, we select DDIM, a deterministic diffusion model as our decoder. To integrate $z_{sem}$ into DDIM, we adopt adaptive Group Normalization(AdaGN) similar to ~\cite{dhariwal2021diffusion, bao2023unidiffuser} by applying channel-wise scaling ~$\&$ shifting twice to the normalized feature map $\bm{h} \in \mathbb{R}^{c\times h \times w}$.
\begin{equation}
    \text{AdaGN}(\bm{h}, t, z_{sem}) = z_s (t_{s} \text{GroupNorm}(\bm{h}) + t_b)+z_b
\end{equation}
where $z_s, z_b\in\mathbb{R}^c\!=\!\text{MLP}(z_{sem})$ and $t_s, t_b\in\mathbb{R}^c\!=\!\text{MLP}(\psi(t)),\psi$ is used for creating sinusoidal timestep embeddings.

We reimplement the conditional DDIM based on the architecture of a Pandey et al.~\cite{pandey2022diffusevae} and still adopt the loss function \Eqref{equ:l_simple} in DDPM ~\cite{ho2020denoising} to optimize our whole diffusion autoencoder.
\subsection{Refining the latent space} 
After completing the pre-training, we aim to refine the latent space learned by Encoder~$\boldsymbol{E}_\phi$, where each part of the refined latent code represents an attribute. Typically, the refined latent code $\bm{z}_{dis}$ is equally divided into the number of dataset attributes. Inspired by GZS-Net~\cite{GeAXI21}, we propose Group-supervised AutoEncoder(GAE) trained on $z_{sem}$ dataset $\mathcal{D}_z$ in a \textbf{GROUP} manner. Unlike GZS-Net, our proposed GAE is trained on \textbf{GROUP} latent codes, compared to the pixel-wise GZS-Net, our latent-wise approach can achieve faster training and inference speed.

Specifically, we give our definition on \textbf{GROUP}$\mathbf{(C, s)}$.
\begin{definition}
\label{label:group}
\rm{\textbf{GROUP}}$\mathbf{(C, s)}$: Given a dataset $\mathcal{D}$ containing $n$ samples $\mathcal{D} = \{x_{i}\}_{i=1}^n$, each sample $x_i$ is made up of $m$ attributes. We begin by selecting a sample $x_C = \{x_C^{(1)}, x_C^{(2)}, \cdots, x_C^{(m)}\}$.
We then form a set $\mathcal{A}$ by selecting $m$ attributes from $x_C$, resulting in a set of $C_m^s$ elements
Finally, we select $C^s_m$ samples from $\mathcal{D}$ that correspond to $C^s_m$ elements in $\mathcal{A}$. Formally:
\begin{equation}
    \rm{\textbf{GROUP}} \it\mathbf{(C, s)} \Leftrightarrow x_C \cup \{x_i \mid \{x_i^{(\rm 1)}, \it x_i^{(\mathbf{e})}, x_C^{(m)}\} ~~ for ~\mathbf{e} ~in ~\mathcal{A} \}
\end{equation}
\end{definition}

\begin{figure}[t]
\centering
\includegraphics[width=1.\linewidth]{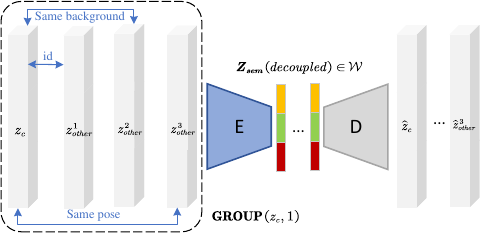}
\caption{GROUP$(z_c, 1)$ of~$z_{sem} ~\text{dataset} ~\mathcal{D}_z$ on ilab\_20M. 
Our Group AutoEncoder(GAE) is a common encoder-decoder architecture. $z_{sem}(decoupled)$ is equivalent to $z_{dis}$ in our paper.} 
\label{fig:group_overview}
\end{figure}

We select a sample $x_C$ from the dataset $\mathcal{D}$ to generate GROUP$(x_C,1)$. By iterating this process and passing the result data to encoder of the aforementioned diffusion autoencoder, we obtain dataset $\mathcal{D}_z^1$ composed of latent code in a group manner as our training dataset for GAE. We leave the case of $i>1$ for future study. In each group of dataset $\mathcal{D}_z^1$, there is only one same attribute value between $z_c$ and each of remaining $z_{sem}$, \ie ~$|\mathbf{A}(z_c) \cap \mathbf{A}(z_{other})|=1$, where $z_{other}$ is every $z_{sem}$ in a group except $z_c$ and $|\mathbf{A}(z)|$ is used for calculating how many attributes in $z$. \Figref{fig:group_overview} illustrates the structure of our group in $\mathcal{D}_z$. 
To further disentangle $z_{sem}$ produced by diffusion autoencoders, we utilize supervised attribute-swap and unsupervised attribute-swap strategies based on $\mathcal{D}_z^1$ in the training process of GAE.

\noindent \textbf{Supervised Attribute-Swap}
Similar to GZS-Net~\cite{GeAXI21}, we can consider the attribute of $z_{sem}$ to be equivalent to the image attribute because there exists a deterministic mapping between them. It is straightforward to consider that two $z_{sem}$ with identical values for one attribute should have near-exact reconstructions when we swap the part of $z_{sem}(decoupled)$ which represents the aforementioned attribute. In addition, since each of the $z_{other}$ has and only has one attribute value identical to $z_c$, we reassemble the parts of each $z_{other}$ that represent the same properties as $z_c$ and get $z_{re}$. The reconstructed $\hat{z}$ from all $z_{re}$ should also be almost identical to $z_c$. We can encourage disentanglement in the latent space of attributes in this way and obtain the supervised attribute-swap loss $\mathcal{L}_{ss}$

\noindent \textbf{Unsupervised Attribute-Reassemble}
We must impose stricter constraints on the GAE to perform zero-shot image synthesis. \ie we can reconstruct the image by GAE even if it does not exist in the dataset. So we combine the above idea of regression matching by swapping latent variable representations and incorporate the idea of Cycle-GAN~\cite{zhu2017unpaired}. We select one part from each $z_{sem}(decoupled)$ and reassemble them into $z_{u}$. $\hat{z}$ generated by the decoder of GAE from $z_u$ will re-enter into the same GAE for regression matching between $z_u$ and $\hat{z_u}$. Formally, we have:
\begin{equation}
    z_{sem} \stackrel{\textbf{E}}{\longrightarrow} z_{dis} \stackrel{f}{\longrightarrow} z_u \stackrel{\textbf{D}}{\longrightarrow} \hat{z} \stackrel{\textbf{E}}{\longrightarrow} \hat{z_u}
\end{equation}
where \textbf{E} for encode process, \textbf{D} for decode process, $f$ for the reassemble operation respectively. By calculating the $L1$ loss between $z_u$ and $\hat{z}_u$, then we obtain its unsupervised attribute-reassemble loss $\mathcal{L}_{ur}$.

We combine a total loss $\mathcal{L}_{dis}$ to disentangle the latent space generated by diffusion autoencoder. Formally, we have:
\begin{equation}
    \mathcal{L}_{dis} = \mathcal{L}_r + \lambda_{ss}\mathcal{L}_{ss} + \lambda_{ur}\mathcal{L}_{ur}
\end{equation}
where $\mathcal{L}_r, \mathcal{L}_{ss}, \mathcal{L}_{ur}$ respectively represents for reconstruction, supervised attribute swap, and unsupervised attribute reassemble losses. $\lambda_{ss},\lambda_{ur}>0$ is scalar hyperparameters. 
Our suggestion values for $\lambda_{ss}$, $\lambda_{ur}$ are 1.0 and 0.5, respectively. We train the whole GAE via Adam~\cite{KingmaB14adam} optimizer with learning rate of 4e-5 and no weight decay.

\noindent \textbf{Fine-tuning via jointly training}
The gap between $\hat{z}_{sem}$ generated by GAE and $z_{sem}$ extracted from the encoder of the diffusion autoencoder is unavoidable. To minimize this gap, we need to train them in an end-to-end manner. Since our GAE is designed as a plug-in module, we can directly connect it to the diffusion autoencoder as shown in Fig.\ref{fig:pipeline}. We integrate the loss of diffusion autoencoder and GAE together as our final loss $\mathcal{L}$. Formally:
\begin{equation}
    \mathcal{L} = \mathcal{L}_{simple} + \bm{\gamma}\mathcal{L}_{dis}
\end{equation}
where $\mathcal{L}_{simple}$ is loss for diffusion autoencoder and $\mathcal{L}_{dis}$ is for GAE. $\bm\gamma$ control the degree of disentanglement of latent space,~default 1. $\lambda_{ss}$ and $\lambda{ur}$ are set respectively 0.5 and 0.5. We adopt the same Adam optimizer with no weight decay and learning rate of 1e-5.
Each attribute is fixed to a part of $z_{dis}$ when we finish training, we can achieve multi-attribute image manipulation by replacing the corresponding part in $z_{dis}$.

Our choice of training GAE post-hoc has a few empirical reasons. Firstly, We found it unstable when we train our diffusion autoencoder and GAE in an end-to-end manner. We attribute it to the gap of different optimation directions between them. So we take this way to ensure the powerful generation capability of diffusion autoencoder. Secondly, since our GAE is trained on low-dimension latent code, GAE training takes only a fraction of the whole training time, enabling quick experiments on different GAE with the same diffusion autoencoder.
\vspace{-8pt}
\subsection{Group AutoEncoder(GAE) Design}
\Figref{fig:group_overview} shows the overview of GAE and how it works based on GROUP $z_{sem}$. Specifically, similar to all kinds of autoencoders, 
our GAE is composed of an encoder and decoder as usual, both encoder and decoder are made up of a few MLP blocks and bottleneck blocks as Fig.~\ref{fig:pipeline}.b. Specifically, there are four MLPs and two BottleNecks in both encoder and decoder
Regarding the selection of norm layer, we adopt LayerNorm and GroupNorm in blocks of encoder and decoder depending on different properties of their input $\bm{z}$. We choose Silu as our activation function in all blocks.

We experiment with multiple architectures for the two blocks including MLP, MLP + skip connection, and using 2-D feature map from the last stage of encoder $E_{\phi}$ as input to GAE.
We have found that MLP + skip connection performs better while being very effective. In addition, we empirically find that the last block of GAE has to be skip connection, otherwise, training of the model will be unstable.

\section{Experiments}
\label{sec:experiments}
We evaluate our model on ilab\_20M~\cite{borji2016ilab}, a collection of various kinds of vehicle images captured from different angles, with diverse attributes and backgrounds. We adopt the truncated version of ilab\_20M which comprises three attributes: identity(6), background(111), and pose(10). Our aim is to leverage our DiffuseGAE to conduct image manipulation in a higher-level semantic latent space.

Following PDAE~\cite{zhang2022unsupervised} and GZS-Net~\cite{GeAXI21}, we respectively set dimensions of $\bm{z}_{sem}$ and $\bm{z}_{dis}$ to 512 and 100 for all datasets. For ilab\_20M, we partition the latent code among attributes as 60 for identity, 20 for background, and 20 for pose.

\begin{table*}[htb]
\caption{Perplexity matrix on different models and diagonals are bolded. We trained nine linear classifiers to compute the perplexity matrix for disentangled representation analysis. For AE + DS(directly supervised), we adopt the same autoencoder as GZS-Net, but We directly apply the category ground-truth as supervisory information to constrain the latent space generated by the autoencoder. Thus, it is reasonable to expect better performance of AE + DS in the perplexity matrix evaluation. The attribute partition in the latent code is consistent with our method.}
\label{tab:c_matrix}
\renewcommand{\arraystretch}{1.2}
\centering
\footnotesize
\setlength\tabcolsep{5pt}
  \begin{tabular}{c|ccc|ccc|ccc|ccc}
    \toprule
    \multirow{2}{*}{$\mathcal{A}$ ($|\mathcal{A}|$)}& \multicolumn{3}{c|}{\textbf{\textit{DiffuseGAE(Ours)}}} & \multicolumn{3}{c|}{\textit{Diff-AE}} & \multicolumn{3}{c|}{\textit{GZS-Net}} & \multicolumn{3}{c}{\textit{AE} + \textit{DS}} \\
     \cline{2-13}
    & \text{Identity} & \text{Background} & \text{Pose}
    & \text{Identity} & \text{Background} & \text{Pose}
    & \text{Identity} & \text{Background} & \text{Pose}
    & \text{Identity} & \text{Background} & \text{Pose} 
    \\

    \midrule
    \text{Identity} (6)
    & \bf{.96} & .019 & .168 
    & \bf{.90} & .018 & .167 
    & \bf{.93} & .020 & .17 
    & \bf{.97} & .015 & .16 
    \\
    \text{Background} (111) 
    & .15 &  \bf{1.0} & .14 
    & .132 & \bf{.91} & .167 
    & .133 & \bf{1.0} & .167 
    & .14  & \bf{1.0} & .142 
    \\
    \text{Pose}(10)
    & .12 & .015 & \bf{.99} 
    & .10 & .019 & \bf{.93} 
    & .11 & .019 & \bf{.97} 
    & .14 & .016 & \bf{.99}
    \\
    \bottomrule
  \end{tabular}
\end{table*}

\subsection{Disentanglement results on latent space}

We qualitatively and quantitatively evaluate our method on its ability to decouple the latent space $z_{sem}$. 

\begin{figure}[htb]
\centering
\vspace{-2pt}
\includegraphics[width=0.85\linewidth]{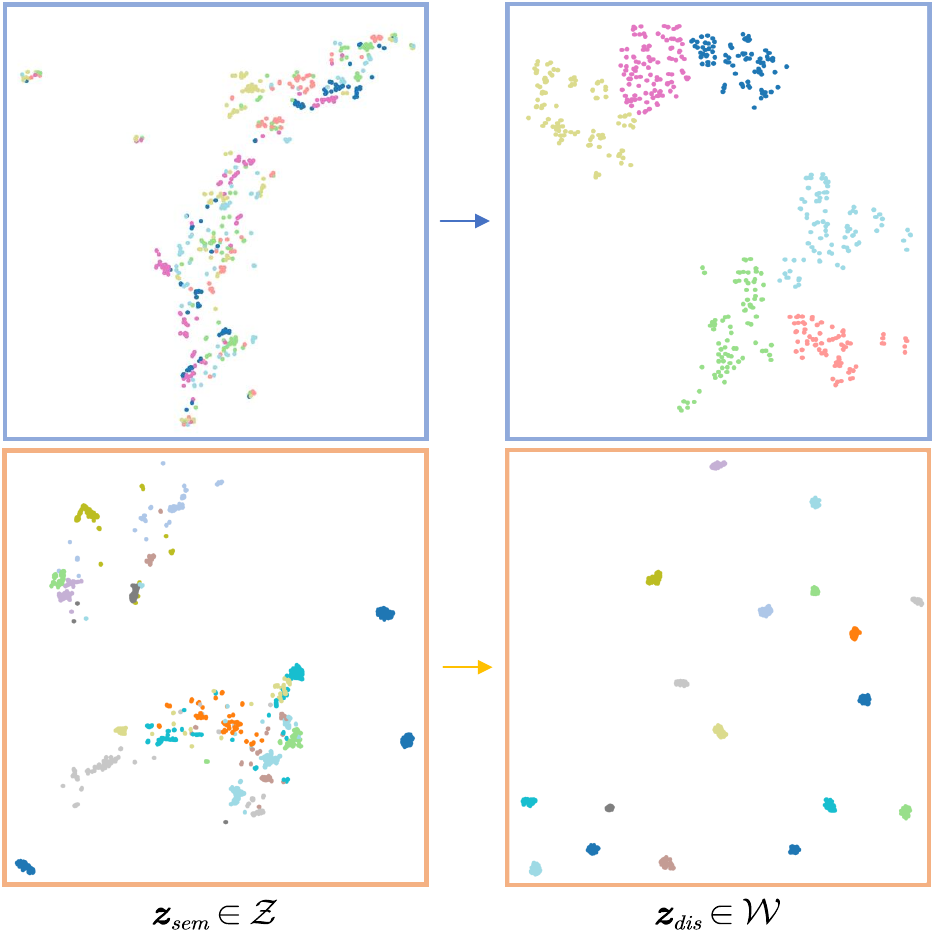}
\caption{T-SNE visualization on the latent codes $\bm{z}_{sem}$ and $\bm{z}_{dis}$ of ilab\_20M dataset, representing semantic and disentangled information, respectively. The samples belonging to different identities and backgrounds are represented in blue and orange, respectively. For clarity, we only select 6 attribute values for identity and 18 for background.}
\label{fig:tsne}
\end{figure}

We visualize the latent space of the ilab\_20M dataset using T-SNE to gain insights into the dataset and identify any patterns or clusters based on identity and background attributes. The results of our proposed GAE are presented in Fig. \ref{fig:tsne}, which clearly demonstrate the effectiveness of our method in exploiting the latent space $\mathcal{Z}$ and effectively decoupling $z_{sem}$.

Under perfect disentanglement, the latent code of one attribute should always predict its attribute value. The better disentanglement we get, the more accurate the prediction will be when we use the corresponding part in $z_{sem}$. Thus we train a linear classifier for each of the attributes and use it to calculate perplexity matrix among attribute pairs. We denote ~$|\mathcal{A}_r|$ as the number of attribute values in attribute $r$. For all datasets, we splits them 80:20 for $\mathcal{D}_{train} ~\& ~\mathcal{D}_{test}$. Table \ref{tab:c_matrix} shows the results on three attributes of ilab\_20M. the best case should be 1 on the main diagonal and random values,\textit{i.e.} 1/$|\mathcal{A}_r|$ on the off-diagonal. DiffuseGAE outperforms baseline and other disentanglement methods except for AE + DS as it is trained by imposing such constraints on the latent space directly. Therefore it is reasonable that AE + DS performs better in the perplexity matrix evaluation. But our DiffuseGAE achieves comparable results in this metric with AE + DS and it shows inferior image synthesis performance than other methods including ours shown in \twoFigref{fig:img_replacement}{fig:img_recombination}.

\subsection{Performance on zero-shot image synthesis}
We conduct two experiments to evaluate our DiffuseGAE in zero-shot image synthesis. \textit{i.e.} image replacement and image recombination.

\begin{figure}[htb]
\centering
\vspace{-5pt}
\includegraphics[width=1\linewidth]{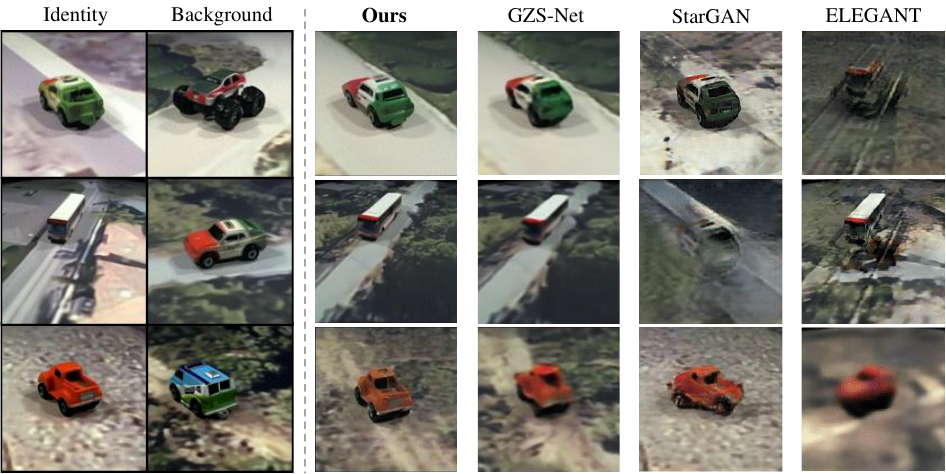}
\caption{
Image replacement performance on ilab\_20M. Columns right of the dashed line represent the output of various methods, the left shows two input baselines for image synthesis.
StarGAN~\cite{choi2018stargan} receives one input image and \textit{target-domain information} to achieve image manipulation, While ELEGANT~\cite{xiao2018elegant} takes
identity and background images as inputs to perform image synthesis.
}
\label{fig:img_replacement}
\end{figure}

\Figref{fig:img_replacement} presents the qualitative results of our approach on ilab\_20M. We compare ours with two methods based on autoencoding, GZS-Net and AE + DS, and against two disentangled and controllable GAN baselines: StarGAN~\cite{choi2018stargan} and ELEGANT~\cite{xiao2018elegant}. Our method performs image replacement as follows: 
We choose two images, the former provides the part in ~$\bm{z}_{dis}$ corresponding to attribute id and pose, while the latter provides the part corresponding to attribute background. We follow Eq.(\ref{equ:reversed_process}) to generate noise map~$x_T$ from column identity images. Then we pass the $\hat{\bm{z}}_{sem}$ decoded from recombined $\hat{\bm{z}}_{dis}$ as condition to generate images with ~$T$ = 100 steps. We can clearly see that our method can effectively improve the quality of generation while ensuring the semantics remains unchanged.

\begin{figure}[tb]
\centering
\includegraphics[width=0.95\linewidth]{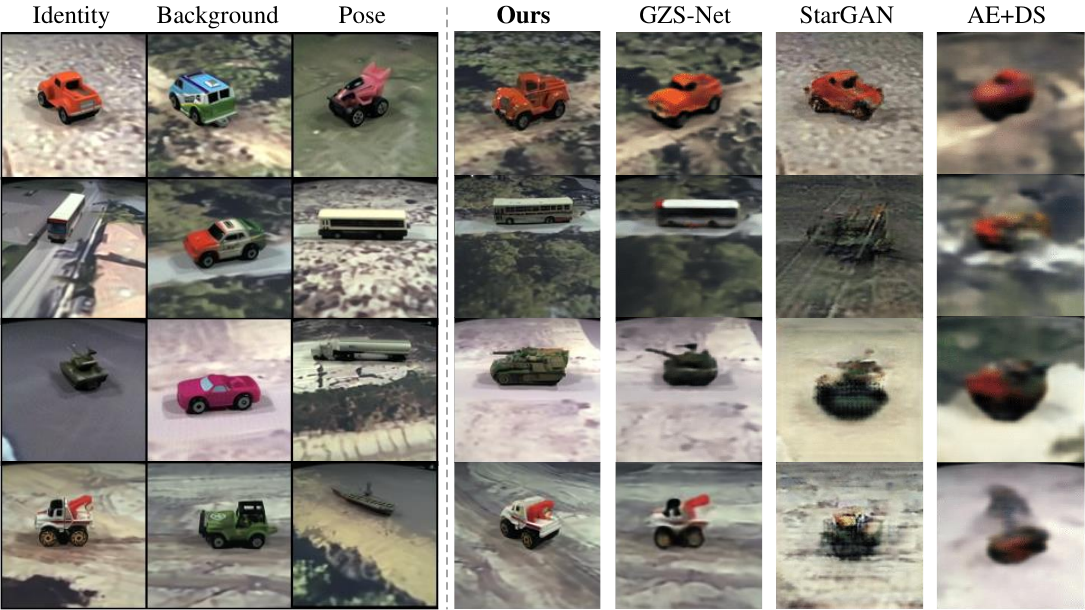}
\caption{Image recombination results on ilab\_20M. We sample ~$x_{T} \sim \mathcal{N}(0, \mathbf{I})$ as noise map with $T$ =100 steps. Similar to image replacement, we use the left three of dashed line to generate conditions for image generation.
}
\label{fig:img_recombination}
\vspace{-5pt}
\end{figure}

For image recombination, We compare it against GZS-Net, AE + DS, and StarGAN~\cite{choi2018stargan} in image recombination. Similar to image replacement, we select three images to provide attribute identity background and pose. Each image is responsible for providing an attribute. Then we reassemble them into $\hat{z}_{dis}$ and pass their decoding $z_{sem}$ to conditional DDIM as condition. Since there is no such target image we wish to synthesize, we adopt $x_T \sim \mathcal{N}(0, \textbf{I})$ as noise map to generate images iteratively. Our DiffuseGAE is significantly better than other models in terms of generation quality. And other models which can be used for manipulating attributes are often black-box models, \textit{i.e.} strong in knowing what to change but not in how to change. Yet our method has a more intuitive pipeline for attribute manipulation.
\Figref{fig:img_recombination} shows our generative results on image recombination. 

\vspace{-5pt}
\subsection{Autoencoding reconstruction quality}
Although we aim to obtain a better decoupled latent space and good reconstruction quality of an autoencoder might not be a good indicator of disentangled representation learning, it is worth noting that good reconstruction quality in autoencoders is often indicative of better information compression and can be useful for image manipulation.

To quantitatively analyze performance on disentanglement of latent code, We randomly select 50k \textbf{GROUP}$\mathbf{(C, 1)}$ images as $\mathcal{D}_g$ for evaluation. 

We get $z_{dis}$ of all $x_{other}$ via DiffuseGAE in each group and reassemble them according to their corresponding attribute. We take the reassembled latent code ~$\hat{z}_{dis}$ to generate 50k images for fid evaluation. We compute FID($\downarrow$) between the 50k generated images and ~$x_C$ in each group (50k in total).
In order to fully assess the effectiveness of the reconstruction, we also evaluated methods on a wider range of criteria, \textit{i.e.} SSIM($\uparrow$)~\cite{wang2003multiscale}, LPIPS($\downarrow$)~\cite{zhang2018unreasonable}, MSE($\downarrow$) in Table ~\ref{table:autoencoding}. As we can see, DiffuseGAE outperforms other methods under similar latent dimensions. 

\begin{table}
\renewcommand{\arraystretch}{1.25}
    \caption{Autoencoding reconstruction quality of models trained on ilab\_20M. The quality of reconstruction is evaluated using various metrics on 50,000 pairs of images. It is noteworthy that the latent dimension of the models in the first part is smaller than those in the second part, specifically 100, 512, 100 and 49152, 49252, respectively.}
    \label{table:autoencoding}
    \centering
    \footnotesize
    \setlength{\tabcolsep}{2pt}
    \begin{tabular}{l|cccc}
    \toprule
    
    \textbf{Model} & \textbf{FID} $\downarrow$ &  \textbf{SSIM} $\uparrow$ &\textbf{LPIPS} $\downarrow$ & \textbf{MSE} $\downarrow$ \\
    \midrule
    GZS-Net (No T and ~$x_T$)~\cite{GeAXI21}                             & 100.85 & 0.345 & 0.205 & 0.124 \\
    Diff-AE  (T=100, random $\bm{x}_{T}$)~\cite{preechakul2022diffusion} & 8.07 & 0.539 & \textbf{0.073} & \textbf{0.011} \\
    \rowcolor{gray!20}
    \textbf{DiffuseGAE(T=100, random $\bm{x}_{T}$)}                      & \textbf{8.06} & \textbf{0.549} & \textbf{0.073} & \textbf{0.011} \\
    \hline
    
    DDIM  (T=100)~\cite{song2020denoising}                                & 10.37 & 0.912 & 0.063 & 0.002 \\
    \rowcolor{gray!20}
    \textbf{DiffuseGAE(T=100, inferred $\bm{x}_{T}$)}                     & \textbf{4.03} & \textbf{0.989} & \textbf{0.012} & \textbf{8.56e-5} \\
    \bottomrule
    \end{tabular}
\vspace{-10pt}
\end{table}
\subsection{Interpolation of meaningful latent code and trajectories}
Given two images $x_0^{(1)}$ and ~$x_0^{(2)}$ from dataset~$\mathcal{D}$, We encode them into $(z_{dis}^{(1)}, ~x_T^{(1)})$ and ~$(z_{dis}^{(2)}, ~x_T^{(2)})$ using our trained model. We evaluate our method qualitatively and quantitatively by running our generative process of conditional DDIM from ~$Slerp(x_T^{(1)}, x_T^{(2)}; \alpha)$ under the condition $\hat{z}_{sem}$ with 100 steps, expecting smooth trajectories in the direction along $\alpha$. The $\hat{z}_{sem}$ is decoded from $Lerp(z_{dis}^{(1)}, z_{dis}^{(2)}; \alpha)$ by GAE. 
\begin{figure}[htb]
\centering
\includegraphics[width=1\linewidth]{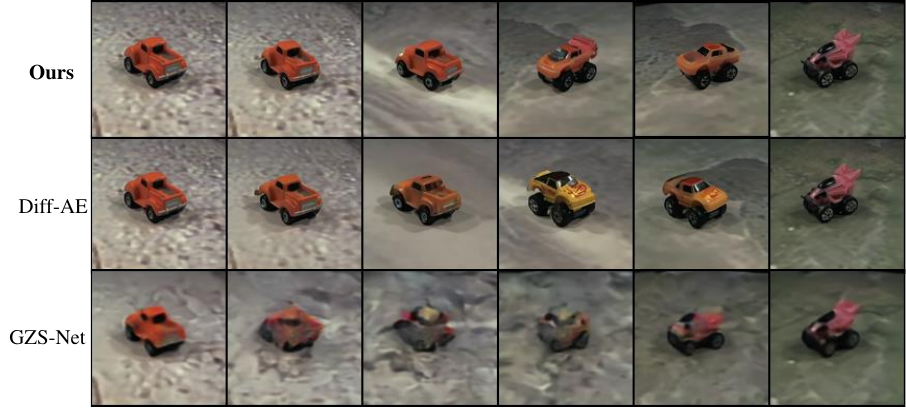}
\caption{Multi-attribute image interpolation performance on ilab\_20M. The real images are displayed in the first and last columns. The first two rows show interpolation for identity only, while the third and fourth rows represent interpolation for background and pose, respectively.}
\label{fig:img_interpolation_multi}
\vspace{-5pt}
\end{figure}

\begin{figure*}[t]
\centering
\vspace{-10pt}
\includegraphics[width=0.9\linewidth]{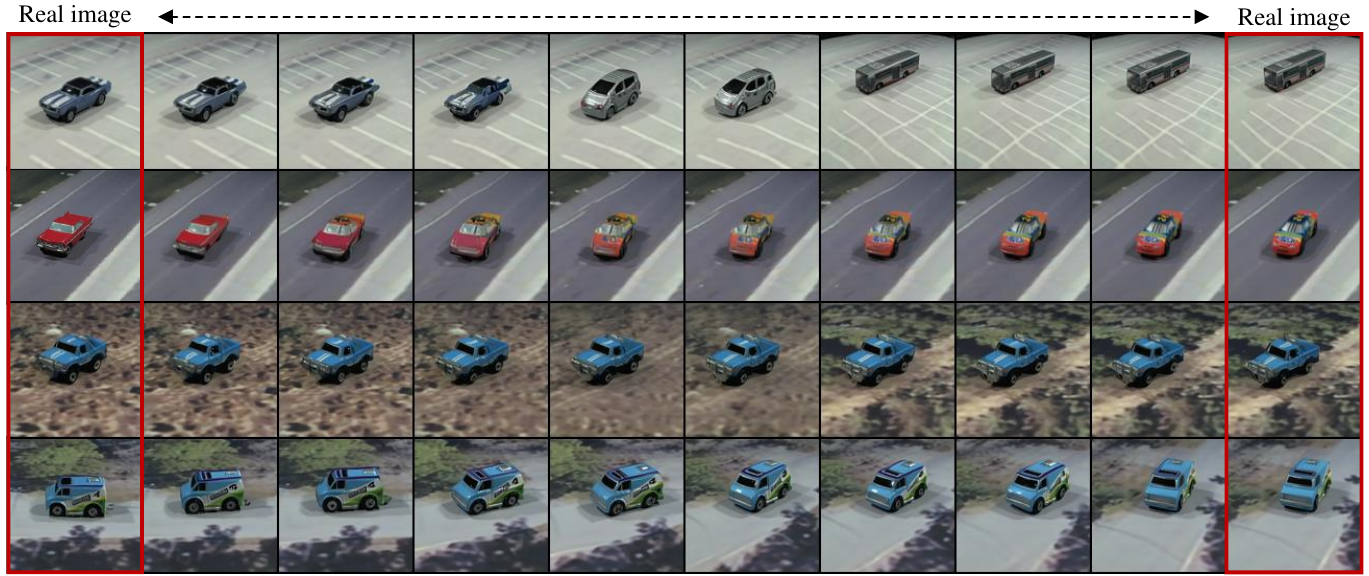}
\caption{Single-attribute image interpolation performance on ilab\_20M. The first and last columns are real images. The first two rows only interpolate for identity, 3rd and 4th rows represent background and pose respectively.}
\label{fig:img_interpolation_single}
\end{figure*}

To evaluate the smoothness of our method for single-attribute interpolation. We present examples of each attribute in \Figref{fig:img_interpolation_single}. We only work on the part of ~$z_{dis}$ that corresponds to the attribute we desire to interpolate. For attribute \textit{identity}, we respectively show our results on inter-class and intra-class interpolation. \Figref{fig:img_interpolation_single} shows the trajectories for various values of ~$\alpha \in [0,1]$ among three attributes. Our methods produce smooth results with well-preserved details from both images and also hold robust semantic consistency, \textit{i.e.} only changing what we want, which verifies the disentanglement performance of our approach.
We have also experimented with multi-attribute interpolation on two images where all attributes are different. \Figref{fig:img_interpolation_multi} shows that our method performs better in multi-attribute image interpolation owing to the excellent decoupling capability.

\begin{figure}[h!]
\centering
\setlength{\abovecaptionskip}{-5pt}

\includegraphics[width=0.7\linewidth]{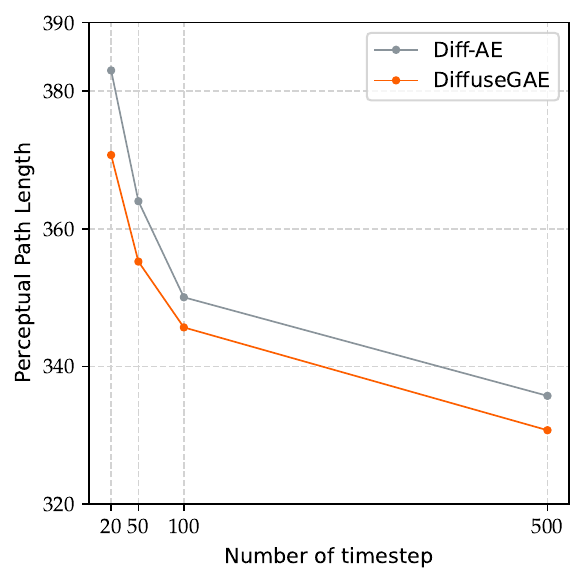}
\caption{Perceptual path length~$(\downarrow)$ for various timesteps on ilab\_20M.}
\label{fig:ppl_distance}
\vspace{-5pt}
\end{figure}
To quantify the smoothness of the interpolation, we adopt Perceptual Path Length(PPL) introduced in StyleGAN~\cite{karras2021alias}. Perfect smoothness means that the distance of the images generated by two $z_{sem}$ close to each other should be small. In particular, we compute the following expectation over 1k pairs of sampled latent codes $(z_1, z_2)$ and ~$\alpha \sim U(0,1)$, Formally:
\begin{equation}
\textbf{PPL}=\mathbb{E}\left[\frac{1}{\epsilon^2} d\left(G\left(\operatorname{Slerp}\left(z_1, z_2 ; \alpha\right)\right), G\left(\operatorname{Slerp}\left(z_1, z_2 ; \alpha+\epsilon\right)\right)\right)\right]
\end{equation}

where G indicates image generator, d computes the distance between two generated images based on VGG16 and $\text{Slerp}(\cdot)$ denotes spherical interpolation.
We compute the metric on randomly selected 1k pairs of images from ilab\_20M with $x_T \sim \mathcal{N}(0, \textbf{I})$. As shown in Fig.~\ref{fig:ppl_distance}, our method outperforms Diff-AE under all timesteps in terms of interpolation smoothness.

\section{Related Work}
\label{sec:related}
\textbf{Disentangled Representation Learning} techniques infer independent latent factors from visual objects on the basis of the hypothesis that each factor contributes to the generation of a single semantic attribute. Following Variational Autoencoders~\cite{kingma2013auto} due to its prior assumption, a class of model~\cite{LiCPZ19, vahdat2020nvae} achieve disentanglement by reducing the KL-divergence between latent code and Norm distribution~$\mathcal{N}(0,\textbf{I})$ which is statistically-independent. while these methods can not achieve ideal disentanglement due to the presence of posterior collapse~\cite{van2017neural} and prior hole~\cite{sinha2021d2c}. 
On the other hand, they are unable to synthesize novel images without seeing before. Stronger constraints need to be imposed on the latent code to obtain combinatorial generalizability. So we make the disentanglement further explicit by imposing an inverse constraint on the original conversion, \textit{i.e.} Supervised Attribute-Swap and Unsupervised Attribute-Reassemble.

\noindent\textbf{Image Manipulation} has always been a challenge in the computer vision community with various practical applications~\cite{karras2021alias, patashnik2021styleclip} and Generative Adversarial Networks (GANs) have attracted widespread attention due to their outstanding performance. StarGAN~\cite{choi2018stargan} jointly train domain information together with images and add mask vectors to facilitate training among different datasets. ELEGANT~\cite{xiao2018elegant} achieves representation disentanglement by constraining semantic encoding positions, thus enabling multi-attribute image manipulation. A wide range of methods~\cite{abdal2020image2stylegan++, richardson2021encoding} is implemented to obtain the mapping from images to latent space, thus semantic and natural modifications can be achieved by editing in the latent space. This is what we have achieved in our work.
In recent years, as diffusion models have become increasingly prevalent alternatives, image manipulation based on diffusion models has been intensively investigated. prompt2prompt~\cite{hertz2022prompt} achieve image manipulation through the cross-attention mechanism of the diffusion model without any specifications over pixel space.

\noindent\textbf{Diffusion Probabilistic Models}~\cite{ ho2020denoising, song2020denoising} are a family of deep generative models that transform Gaussian noise into original images by a multi-step denoising process. They are closely related to score-based generative models~\cite{song2019generative, song2020score}. Models under this family are now popular for their outstanding sample quality rivaling GANs with more training stability. 
Moreover, DPMs have gained attention from the research community and the public due to their impressive performances in conditional generation tasks, such as text-guided image generation~\cite{hertz2022prompt, saharia2022photorealistic}, image super-resolution~\cite{li2022srdiff, saharia2022image}, image segmentation~\cite{amit2021segdiff, wu2023medsegdiff} and conditional speech synthesis~\cite{chen2020wavegrad}.
However, compared with latent variable generative models, such as VAEs and GANs, DPMs only obtain a sequence of pixel-level corrupted images that lack high-level semantic information. In light of this, Diffusion autoencoders~\cite{preechakul2022diffusion, zhang2022unsupervised} are proposed for representation learning via autoencoding. They are competitive with state-of-the-art models on image reconstruction and numerous downstream tasks. We further exploit the semantic information in diffusion autoencoders to achieve more controllable multi-attribute editing.
Moreover, recent works attempt to improve sampling efficiency which has been categorized into two types: learning-free strategies such as PNDM, DPM-solver~\cite{lu2022dpm} and learning-based strategies such as DiffuseVAE~\cite{pandey2022diffusevae}, Progressive distillation~\cite{salimans2022progressive}. DDIM~\cite{song2020denoising} served as our condition diffusion model can be viewed as first-order case of DPM-solver which can improve sampling speed.

\section{Conclusion \& Discussion}
In conclusion, We further explore the disentanglement capability of Diffusion Autoencoder and its potential to extract higher-level semantics.
Conventional diffusion autoencoders tend to fall short in terms of multi-attribute manipulation. To that end, we propose DiffuseGAE, a novel decoupled representation learning framework capable of better image reconstruction and multi-attribute disentanglement and manipulation. 
A group-supervised learning autoencoder(GAE) trained in latent space is the key to our method, allowing us to perform zero-shot image synthesis using diffusion autoencoder and disentangled representation.
We believe that future work can explore multi-level semantic disentanglement, similar to StyleGAN~\cite{karras2021alias}, to achieve more generic image manipulation.



\bibliography{references}
\end{document}